\documentclass{bmvc2k}

\newcommand{\parskiny}{\vspace{-2mm}}
\newcommand{\seckiny}{\vspace{-2mm}}
\newcommand{\figskiny}{\vspace{-3mm}}

\newcommand{\done}[1]{}

\usepackage{subfigure}
\usepackage{times}
\usepackage{epsfig}
\usepackage{graphicx}
\usepackage{amsmath}
\usepackage{amssymb}
\usepackage{verbatim}
\usepackage{rotating}
\usepackage{float}
\usepackage{graphicx}
\usepackage{amsmath}
\usepackage{bm}
\usepackage{hyperref}
\usepackage{url}
\usepackage{bm}
\usepackage{algorithmic}

\floatstyle{ruled}
\newfloat{algorithm}{tbp}{loa}
\floatname{algorithm}{Algorithm}

\title{Object localization in ImageNet \\ by looking out of the window}

\addauthor{Alexander Vezhnevets}{vezhnick@gmail.com}{1}
\addauthor{Vittorio Ferrari}{vferrari@gmail.com}{1}

\addinstitution{
 University of Edinburgh\\
 Edinburgh, Scotland, UK
}

\runninghead{Vezhnevets and Ferrari}{Looking out of the window}

\begin{document}

\maketitle
\seckiny \seckiny
\begin{abstract}
We propose a method for annotating the location of objects in ImageNet. Traditionally, this is cast as an image window classification problem, where each window is considered independently and scored based on its appearance alone.
Instead, we propose a method which scores each candidate window in the context of all other windows in the image, taking into account their similarity in appearance space as well as their spatial relations in the image plane. 
We devise a fast and exact procedure to optimize our scoring function over all candidate windows in an image, and we learn its parameters using structured output regression.
We demonstrate on 92000 images from ImageNet that this significantly improves localization over recent techniques that score windows in isolation~\cite{uijlings13ijcv,vezhnevets14cvpr}.
\end{abstract}

\seckiny
\seckiny
\section{Introduction}
\label{sec:intro}

The ImageNet database~\cite{deng09cvpr} contains over 14 million images annotated by the class label of the main object they contain. However, only a fraction of them have bounding-box annotations ($10\%$). Automatically annotating object locations in ImageNet is a challenging problem, which has recently drawn attention~\cite{guillaumin14ijcv,guillaumin12cvpr,vezhnevets14cvpr}.
These annotations could be used as training data for problems such as object class detection~\cite{dalal05cvpr}, tracking~\cite{leibe07iccv} and pose estimation~\cite{Andriluka10cvpr}.
Traditionally, object localization is cast as an image window scoring problem, where a scoring function is trained on images with bounding-boxes and applied to ones without.
The image is first decomposed into candidate windows, typically by object proposal generation~\cite{alexe12pami,manen13iccv,uijlings13ijcv,dollar14eccv}.
Each window is then scored by a classifier trained to discriminate instances of the class from other windows~\cite{guillaumin12cvpr, wang13iccv, dalal05cvpr, girshick14cvpr, harzallah09iccv, uijlings13ijcv, Vedaldi09}
or a regressor trained to predict their overlap with the object~\cite{vezhnevets14cvpr,blaschko08eccv,vedaldi09nips,lehmann2011bmvc}. Highly scored windows are finally deemed to contain the object.
In this paradigm, the classifier looks at one window at a time, making a decision based only on that window's appearance.

\begin{figure*}
\begin{center}
    \subfigure{\label{subfig:boyandball}\includegraphics[scale=0.27]{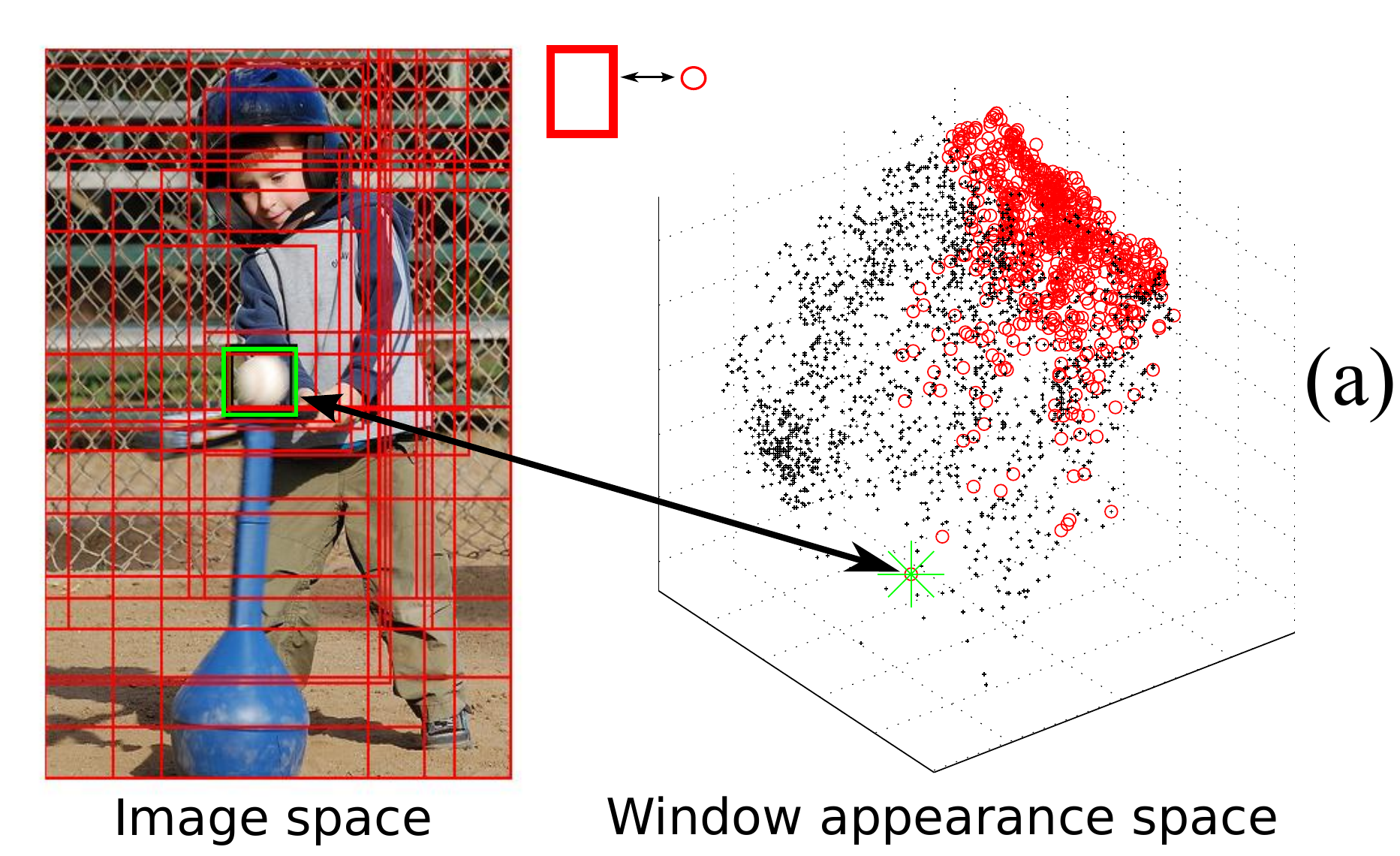}}          
    \subfigure {\label{subfig:wolf}\includegraphics[scale=0.27] {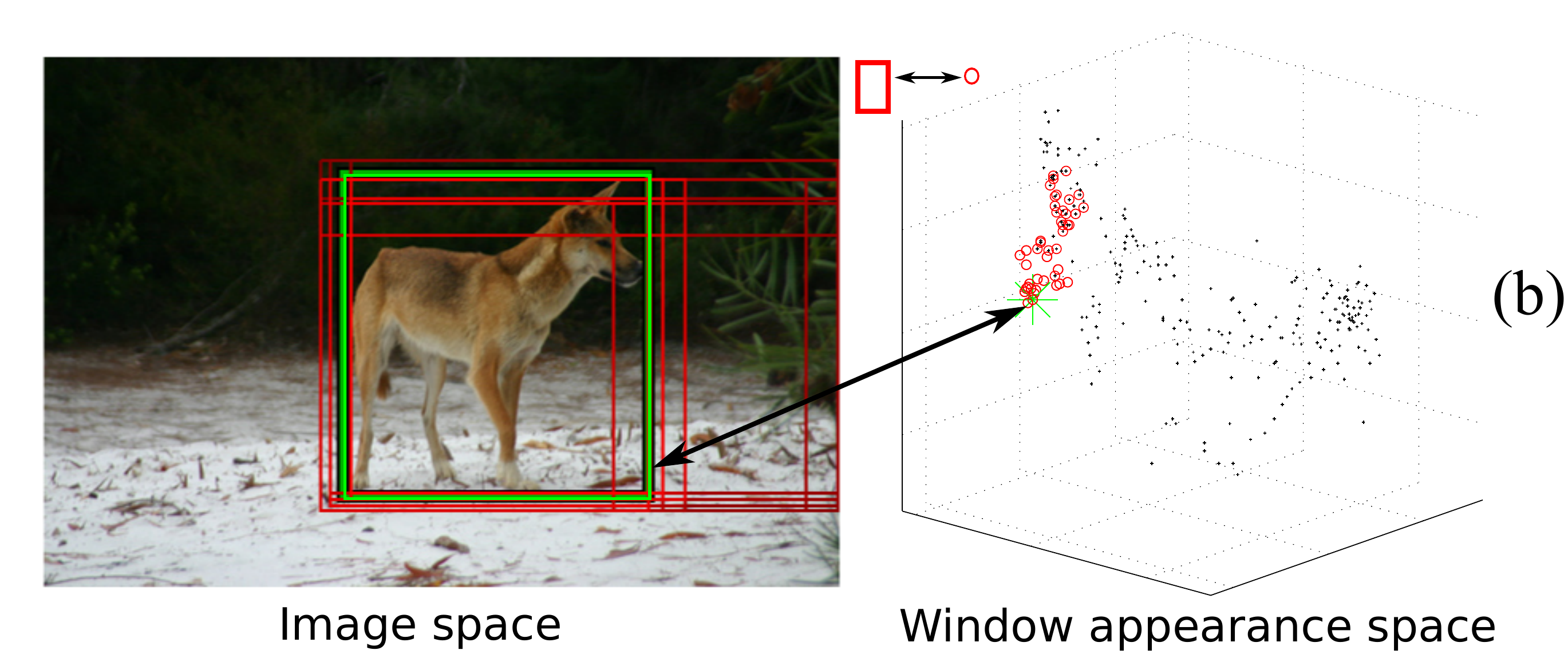} 
} 
\end{center}
\figskiny  
\caption{\small{\it {\bf Connecting the appearance and window position spaces.}
(a) a window tight on the baseball (green star in the appearance space plot) and some larger windows containing it (red circles in the appearance space). Black points in appearance space represent all other candidate windows.
(b) all windows with high overlap with the wolf are shown in red, both in the image and in appearance space. The ground-truth bounding-box to be found is shown in green. 
The appearance space plots are actual datapoints, representing windows in 3-dimensional Associative Embedding of SURF bag-of-words.
Please see main text for discussion.
\label{fig:intuition} \figskiny  \figskiny  }}

\end{figure*}

We believe there is more information in the collection of windows in an image.
By taking into account the appearance of all windows {\em at the same time} and connecting it to their spatial relations in the image plane, we could go beyond what can be done by looking at one window at a time.
Consider the baseball in fig.~\ref{subfig:boyandball}. For a traditional method to succeed, the appearance classifier needs to score the window on the baseball higher than the windows containing it. The container windows cannot help except by scoring lower and be discarded.
By considering one window at a time with a classifier that only tries to predict whether it covers the object tightly, one cannot do much more than that.
The first key element of our work is to predict richer spatial relations between each candidate window and the object to be detected, including part and container relations.
The second key element is to employ these predictions to reason about relations between different windows.
In this example, the container windows are predicted to contain a smaller target object somewhere inside them, 
and thereby actively help by {\em reinforcing} the score of the baseball window.
Fig.~\ref{subfig:wolf} illustrates another example of the benefits of analyzing all windows jointly.
Several windows which have high overlap with each other and with the wolf form a dense cluster in appearance space, making it hard to discriminate the precise bounding-box by its appearance alone. However, the precise bounding-box is positioned at an extreme point of the cluster --- on the tip. By considering the configuration of all the windows in appearance space together we can reinforce its score.

In a nutshell, we propose to localize objects in ImageNet by scoring each candidate window in the context of all other windows in the image, taking into account their similarity in appearance space as well as their spatial relations in the image plane.
To represent spatial relations of windows we propose a descriptor indicative of the part/container relationship of the two windows and of how well aligned they are (sec.~\ref{sec:relation_space}). 
We learn a windows appearance similarity kernel using the recent Associative Embedding technique~\cite{vezhnevets14cvpr} (sec.~\ref{sec:predicting_sr_with_o}).
We describe each window with a set of hyper-features connecting the appearance similarity and spatial relations of that window to all other windows in the same image. These hyper-features are indicative of the object's presence when the appearance of a window alone is not enough (e.g. fig~\ref{fig:intuition}).
These hyper-features are then linearly combined into an overall scoring function (sec.~\ref{sec:the_model}).
We devise a fast and exact procedure to optimize our scoring function over all candidate windows in a test image (sec.~\ref{sec:inference}), and we learn its parameters using structured output regression~\cite{tsochantaridis:jmlr05} (sec.~\ref{sec:str_svm}).

We evaluate our method on a subset of ImageNet containing 219 classes with more than 92000 images~\cite{guillaumin12cvpr,vezhnevets14cvpr,guillaumin14ijcv}. The experiments show that our method outperforms a recent approach for this task~\cite{vezhnevets14cvpr}, an MKL-SVM baseline~\cite{Vedaldi09} based on the same features, and the popular UVA object detector~\cite{uijlings13ijcv}.
The remainder of the paper is organized as follows. Sec.~\ref{sec:relation_space} and~\ref{sec:predicting_sr_with_o} introduce the spatial relation descriptors which we use in sec.~\ref{sec:the_model} to define our new localization model.
In sec.~\ref{sec:related_work} we review related work.
Experiments and conclusions are presented in sec.~\ref{sec:experiments}.

\seckiny
\section{Describing spatial relations between windows}
\label{sec:relation_space}

\parskiny
\paragraph{Candidate windows.}
Recently, object class detectors are moving away from the sliding-window paradigm and operate instead on a relatively small collection of \emph{candidate windows}~\cite{alexe12pami,uijlings13ijcv,girshick14cvpr,guillaumin12cvpr,vezhnevets14cvpr,manen13iccv,dollar14eccv,wang13iccv} (also called `object proposals').
The candidate window generators are designed to respond to objects of any class, and typically just $1000 - 2000$ candidates are sufficient to cover all objects in a cluttered image~\cite{alexe12pami,uijlings13ijcv,manen13iccv,dollar14eccv}.
Given a test image, the object localization task is then to select a candidate window covering an instance of a particular class (e.g. cars).
Following this trend, we generate about 1000 candidate windows $W=\{w\}_{i=1}^N$ using the recent method~\cite{manen13iccv}.

\begin{figure*}
\begin{center}
    \subfigure {\label{subfig:box_rel}\includegraphics[scale=0.14]{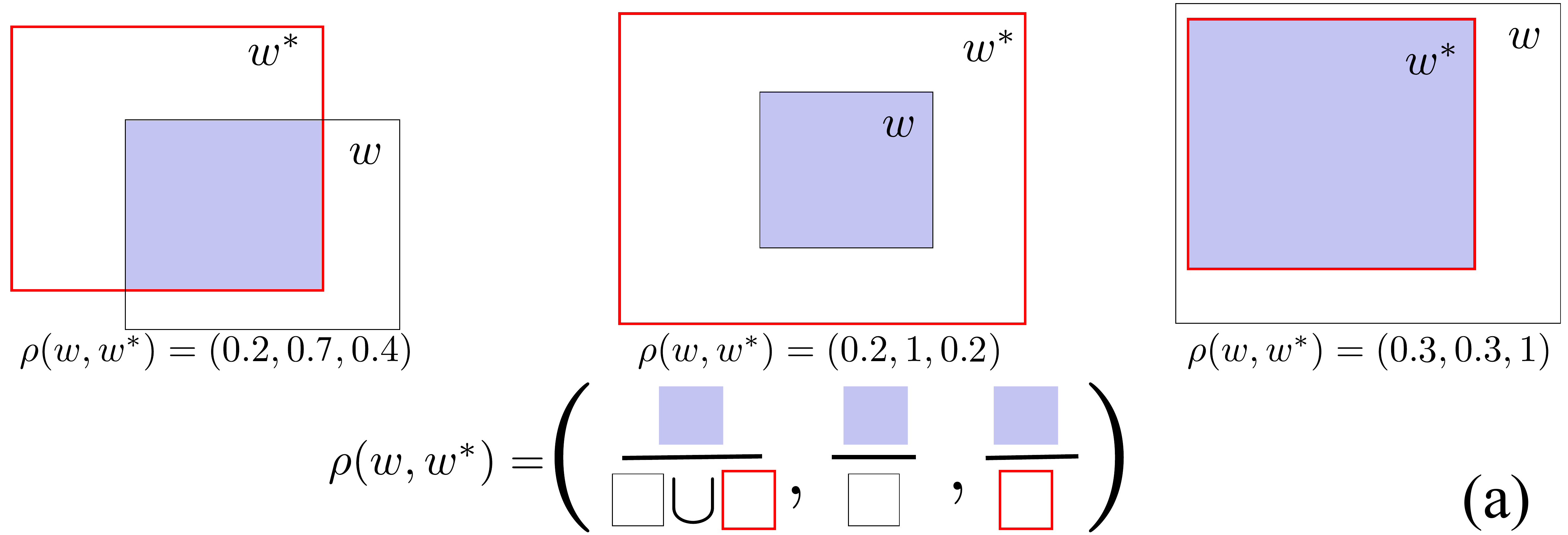}}            \hspace*{22pt}
    \subfigure {\label{subfig:box_yl}\includegraphics[scale=0.23] {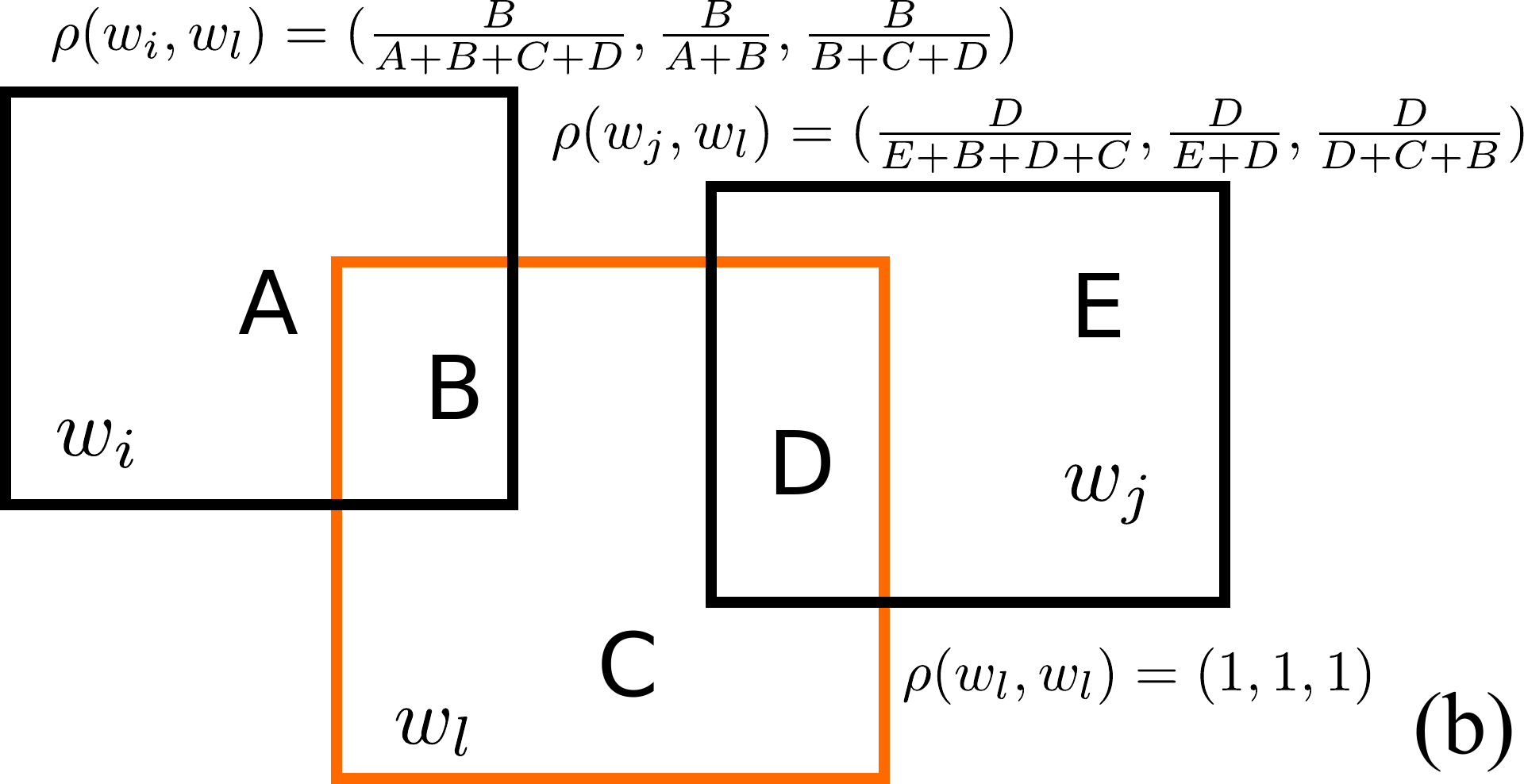}} \hspace*{22pt}       
\end{center}
\figskiny \figskiny \figskiny
\caption{\small{ \em {\bf Spatial relations $\rho(w,w')$ between two windows.}
(a) The first element indicate how much $w$ and $w'$ overlap, in the traditional PASCAL VOC sense~\cite{everingham10ijcv}. The second element indicates whether window $w$ is a part of $w'$. The third element measures whether $w$ is a container of $w'$.
(b) Computing the spatial relations $\rho(w_i,w_l)$ and $\rho(w_j,w_l)$ for hyper-features $\phi_G$ (\ref{eq:spatial_term}) and $\phi_L$ (\ref{eq:spatial_loc_term})} .\figskiny \figskiny}
\label{fig:BoxRelation}

\end{figure*}

\parskiny
\paragraph{Spatial relation descriptor.}
We introduce here a representation of the spatial relations between two windows $w$ and $w'$, which we later use in our localization model (sec.~\ref{sec:the_model}).
We summarize the spatial relation between windows $w$ and $w'$ using the following \emph{spatial relation descriptor} (fig.~\ref{subfig:box_rel})
\begin{equation}
\rho(w,w') = \left( \frac{w \cap w'}{w \cup w'}, \; \frac{w \cap w'}{w}, \; \frac{w \cap w'}{w'} \right)
\end{equation}
where the $\cap$ operator indicates the area of the intersection between the windows, and $\cup$ the area of their union.
The descriptor captures three different kinds of spatial relations.
The first is the familiar intersection-over-union (\emph{overlap}), which is often used to quantify the accuracy of an object detector~\cite{everingham10ijcv,ilsvrc11}. It is $1$ when $w=w'$, and decays rapidly with the misalignment between the two windows. 
The second relation measures how much of $w$ is contained inside $w'$. It is high when $w$ is a {\em part} of $w'$, e.g. when $w'$ is a car and $w$ is a wheel.
The third relation measures how much of $w'$ is contained inside $w$. It is high when $w$ {\em contains} $w'$, e.g. $w'$ is a snooker ball and $w$ is a snooker table.
All three relations are $0$ if $w$ and $w'$ are disjoint and are $1$ if $w$ and $w'$ match perfectly. Hence the descriptor is indicative for part/container relationships of the two windows and of how well aligned they are.

\parskiny
\paragraph{Vector field of window relations.}
Relative to a particular candidate window $w_l$, we can compute the spatial relation descriptor to any window $w$. This induces a vector field $\rho(\cdot,w_l)$ over the continuous space of all possible window positions.
We observe the field only at the discrete set of candidate windows $W$.
A key element of our work is to connect this field of spatial relations to
measurements of appearance similarity between windows.
This connection between position and appearance spaces drives the new components in our localization model (sec.~\ref{sec:the_model}).

\parskiny
\section{Predicting spatial relations with the object}
\label{sec:predicting_sr_with_o}
A particularly interesting case is when $w'$ is the true bounding-box of an object $w^*$. For the images in the training set, we know the spatial relations $\rho(w,w^*)$ between all candidate windows $w$ and the bounding-box $w^*$.
We can use them to learn to predict the spatial relation between candidate windows and the object from window appearance features $x$ in a test image, where ground-truth bounding-boxes are not given.

Following~\cite{vezhnevets14cvpr}, we use Gaussian Processes regression (GP)~\cite{RasmussenWilliams05book} to learn to predict a probability distribution $P(\rho^r(w,w^*)|x) \sim \mathcal{GP}(m(x),K(x,x'))$ for each spatial relation $r \in \{ \mathtt{overlap}, \mathtt{part}, \mathtt{cont} \}$ given window appearance features $x$. We use zero mean $m(x)=0$ and learn the kernel (covariance function) $K(x,x')$ as in~\cite{vezhnevets14cvpr}.
This kernel plays the role of an appearance similarity measure between two windows. 
The GP learns kernel parameters so that the resulting appearance similarity correlates with the spatial relation to be predicted, i.e. so that two windows which have high kernel value also have a similar spatial relation to the ground-truth. We will use the learnt $K(x,x')$ later in sec.~\ref{sec:the_model}.

For a window $w_i$ in a test image, the GP predicts a Gaussian distribution for each relation descriptors. We denote the means of these predictive distributions as $\mu(x_i)=(\mu^\mathtt{overlap}(x_i), \\ \mu^\mathtt{part}(x_i), \mu^\mathtt{cont}(x_i))$, and their standard deviation as $\sigma(x_i)$. The standard deviation is the same for all relations, as we use the same kernel and inducing points.


\seckiny
\section{Object localization with spatial relations}
\label{sec:the_model}

We are given a test image with
(a) set of candidate windows $W=\{w_i\}_{i=1}^N$;
(b) their appearance features $X=\{x_i\}_{i=1}^N$;
(c) the mean $M=\{\mu(x_i)\}_{i=1}^N$ and standard deviation $\Sigma=\{\sigma(x_i)\}_{i=1}^N$ of their spatial relations with the object bounding-box, as predicted by the GP;
(d) the appearance similarity kernel $K(x_i,x_j)$ (sec.~\ref{sec:predicting_sr_with_o}).

Let $w_l\in W$ be a candidate window to be scored.
%
We proceed by defining a set of hyper-features $\Phi(X,W,M,l)$ characterizing $w_l$, and then define our scoring function through them. 

\parskiny
\paragraph{Consistency of predicted \& induced spatial relations $\phi_C$}

\begin{equation}
\phi^r_C(X,W,l) = \max_i | \rho^r(w_i,w_l) - \mu^r(x_i) | 
\label{eq:consist_term}
\end{equation}

\begin{figure}
\begin{center}
\includegraphics[scale=0.29] {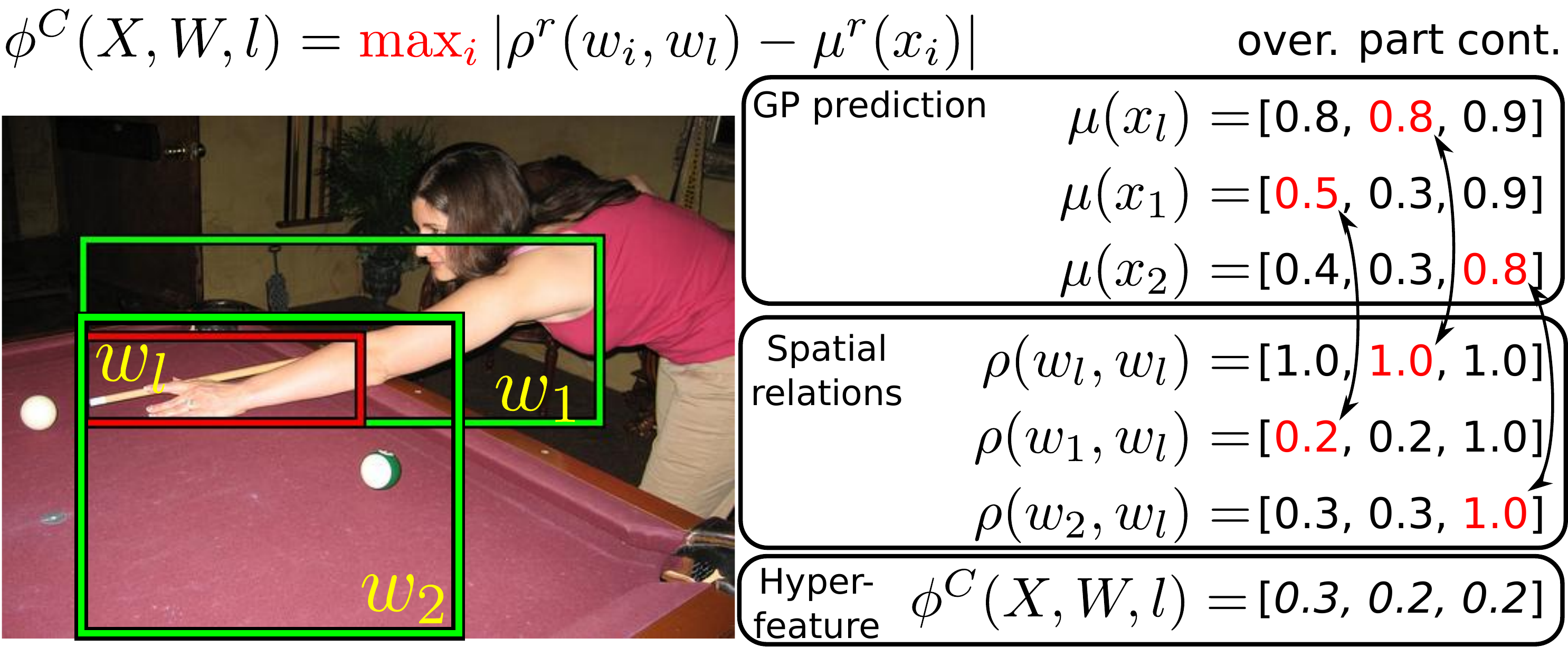} \hspace*{22pt}       
\caption{\small{ \em {\bf Hyper-features $\phi_C$} for a window $w_l$ are computed by finding a maximum disagreement between spatial relations that are predicted by GP $\mu^r(x_i)$ and induced by spatial relations $\rho(w_i,w_l)$ for all other windows $w_i$ in the image. The figure illustrates this on a toy example with three windows $w_l, w_1, w_2$. For each $r \in \{\mathtt{overlap} ,\mathtt{part},\mathtt{cont}\}$ the pairs of $\mu^r(x_i)$ and $\rho(w_i,w_l)$ with maximum disagreement are highlight in red.}\figskiny \figskiny \figskiny} 
\label{fig:StarTreck_ill}
\end{center}
\end{figure}

Assume for a moment that $w_l$ correctly localizes an instance of the object class. Selecting $w_l$ would induce spatial relations $\rho^r(w_i,w_l)$ to all other windows $w_i$.
The hyper-feature $\phi_C$ checks whether these induced spatial relations are consistent with those predicted by GP based on the appearance of the other windows ($\mu^r(x_i)$). If so, that is a good sign that $w_l$ is indeed the correct location of the object.
More precisely, the hyper-feature measures the disagreement between the induced $\rho^r(w_i,w_l)$ and predicted $\mu^r(x_i)$ on the window $w_i$ with the largest disagreement. Fig.~\ref{fig:StarTreck_ill} illustrates it on a toy example. The maximum disagreement, instead of a more intuitive mean, is less influenced by disagreement over background windows, which are usually predicted by GP to have small, but non-zero relations to the object.
It focuses better on the alignment of the peaks of the predicted $\{\mu^r(x_i)\}_{i=1}^N$ and observed $\{\rho^r(w_i,w_l)\}_{i=1}^N$ measurements of the vector field $\rho^r(\cdot,w_l)$, which is more indicative of $w_l$ being a correct localization. 


\parskiny
\paragraph{Global spatial relations \& appearance $\phi_G$}
\begin{equation}
\phi_G^r(X, W,l)={  \frac{2}{N^2-N} \sum_{i=1}^{N-1} \sum_{j=i+1}^N 
| \rho^r(w_i,w_l) - \rho^r(w_j,w_l) | \cdot K(x_i,x_j)}.
\label{eq:spatial_term}
\end{equation}

This hyper-feature reacts to pairs of candidate windows $(w_i,w_j)$ with similar appearance (high $K(x_i,x_j)$) but different spatial relations to $w_l$.
Two windows $w_i,w_j$ contribute significantly to the sum if they look similar (high $K(x_i,x_j)$) {\em and} their spatial relations to $w_l$ are different (high $| \rho^r(w_i,w_l) - \rho^r(w_j,w_l) |)$.

A large value of $\phi_G$ indicates that the vector field of the spatial relations to $w_l$ is not smooth with respect to appearance similarity.
This indicates that $w_l$ has a special role in the structure of spatial and appearance relations within that image.
By measuring this pattern, $\phi_G$ helps the localization algorithm to select a better window, when the information contained in appearance features of $w_l$ alone is not enough.
For example, a window $w_l$ tightly covering a small object such as the baseball in fig.~\ref{subfig:boyandball} has high $\phi_G^{\mathtt{cont}}$, because other windows containing it often look similar to windows not containing it.
In this case, a high value of $\phi_G$ is a positive indication for $w_l$ being a correct localization.
On the other hand, a window $w_l$ tightly covering the wolf in fig.~\ref{subfig:wolf} has low $\phi_G^{\mathtt{overlap}}$, because windows that overlap with it are all similar to each other in appearance space.
In this case, this low value is a positive indication for $w_l$ being correct.
In which direction to use this hyper-feature is left to the learning of its weight in the full scoring function, which is separate for each object class (sec~\ref{sec:str_svm}).

\parskiny
\paragraph{Local spatial relations \& appearance $\phi_L$}

\begin{equation}
{ \phi_{L}^r(X, W,l)= \frac{1}{N} \sum_{i=1}^N | 1 - \rho^r(w_i,w_l) | \cdot K(x_i,x_l)}.
\label{eq:spatial_loc_term}
\end{equation}
This hyper-feature is analogue to $\phi_G$, but focuses around $w_l$ in appearance space. It is indicative of whether windows that look similar to $w_l$ (high $K(x_i,x_l)$) are also similar in position in the image, i.e. their spatial relation $\rho^r(w_i,w_l)$ to $w_l$ is close to $1$.

\parskiny
\paragraph{Window classifier score $\phi_S$.}
The last hyper-feature is the score of a classifier which predicts whether a window $w_l$ covers an instance of the class, based only on its appearance features $x_l$. Standard approaches to object localization typically consider only this cue~\cite{guillaumin12cvpr,wang13iccv,dalal05cvpr,girshick14cvpr,harzallah09iccv,uijlings13ijcv,Vedaldi09}.
In principle, we could use any such method as the score $\phi_S$ here.
In practice, we reuse the GP prediction of the overlap of a window with the object bounding-box as $\phi_s$. One possibility would be to simply use the mean predicted overlap $\mu^\mathtt{overlap}(x_l)$. However, as shown in ~\cite{vezhnevets14cvpr}, it is beneficial to take into account the uncertainty of the predicted overlap, which is also provided by the GP as the standard deviation $\sigma(x_l)$ of the estimate
\begin{equation}
\phi_S(X,l) = [\mu^\mathtt{overlap}(x_l), \;\; \sigma(x_l)]
\label{eq:GP_term}
\end{equation}
Using this hyper-features alone would correspond to the method of~\cite{vezhnevets14cvpr}.

\parskiny
\paragraph{Complete score.}
Let 
\[ 
\Phi(X, W,l) = \left[\{\phi_G^r(X, W,l)\}_{r}, \{\phi_{L}^r(X, W,l)\}_{r}, \{\phi^r_C(X, W,l)\}_{r},\phi_S(X,l)\right]
\]
be a concatenation of all hyper-features defined above for a particular candidate window $w_l$, over all three possible relations: $r \in \{\mathtt{overlap} ,\mathtt{part},\mathtt{cont}\}$. This amounts to 11 features in total.
We formulate the following score function
\begin{equation}
E(\bm{\alpha},X,W,l) = \langle \bm{\alpha}, \Phi(X,W,l) \rangle
\label{eq:the_loss}
\end{equation} 
where $\langle \cdot , \cdot \rangle$ is the scalar product.
Object localization translates to solving \\ $\hat{l} = \arg \max_{l} {E(\bm{\alpha},X,W,l)}$ over all candidate windows in the test image.
The vector of scalars $\bm{\alpha}$ parametrizes the score function by weighting the hyper-features (possibly with a negative weight).
We show how to efficiently maximize $E$ over $l$ in sec.~\ref{sec:inference} and how to learn $\bm{\alpha}$ in sec.~\ref{sec:str_svm}.

\seckiny
\subsection{Fast inference}
\label{sec:inference}

We can maximize the complete score~(\ref{eq:the_loss}) over $l$ simply by evaluating it for all possible $l$ and picking the best one. The most computationally expensive part is the hyper-feature $\phi_G$ (\ref{eq:spatial_term}). For a given $l$, it sums over all pairs of candidate windows, which requires $O(N^2)$ subtractions and multiplications. Thus, a naive maximization over all $l$ costs $O(N^3)$.

To simplify the notation, here we write the score function with only one argument $E(l)$, as the other arguments $\bm{\alpha},X,W$ are fixed during maximization.
Note that $0 \leq | \rho^r(w_i, w_l) - \rho^r(w_j, w_l) | \leq 1$, therefore
\begin{equation}
\alpha^r_G \cdot |\rho^r(w_i, w_l) - \rho^r(w_j, w_l) | \cdot K(x_{i},x_{j})\leq {\left\{ \begin{array}{cc}
0 & ,\alpha^r_G\leq 0\\
\alpha^r_G K(x_{i},x_{j}) & ,\alpha^r_G>0
\end{array}\right\} }
\label{eq:term_bound}
\end{equation}
Where $\alpha_G^r$ is the weight of hyper-feature $\phi_G^r$.
By substituting the elements in the sum over pairs in~(\ref{eq:spatial_term}) with the bound~(\ref{eq:term_bound}), we obtain an upper bound on the $\phi_G^r$ term of the score (in fact three bounds, one for each $r$).
We can then obtain an upper bound $\tilde{E}(l) \geq E(l)$ on the full score by computing all other hyper-features and adding them to the bounds on $\phi_G^r$.
This upper bound  $\tilde{E}(l)$ is fast to compute, as~(\ref{eq:term_bound}) only depends on appearance features $X$, not on $l$, and computing the other hyper-features is linear in $l$.

We use the bound $\tilde{E}$ in an early rejection algorithm.
We form a queue by sorting windows in descending order of $\tilde{E}(l)$.
We then evaluate the full score $E(l)$ of the first $l$ in the queue and store it as the current maximum. We then go to the next $l$ in the queue. If its upper bound $\tilde{E}(l)$ is smaller than the current maximum, then we discard it without computing its full score. Otherwise, we compute $E(l)$ and set it as the current maximum if it is better than the previous best one. We iteratively go through the queue and at the end return the current maximum (which is now the best over all $l$). Notice, that the proposed fast inference method is exact: it outputs the same solution as brute force evaluation of all possible $l$.

%

\seckiny
\subsection{Learning $\alpha$ with structured output regression}
\label{sec:str_svm}

The scoring function (\ref{eq:the_loss}) is parametrized by the weight vector $\bm{\alpha}$. We learn an $\bm{\alpha}$ from the training data of each class using a structured output regression formulation~\cite{blaschko08eccv, Lampert08cvpr,vedaldi09nips,lehmann2011bmvc}. Ideally, we look for $\bm{\alpha}$ so that, for each training image $I$, the candidate window $l_I^*$ that best overlaps with the ground-truth bounding-box has the highest score.
It is also good to encourage the score difference between the best window $l_I^*$ and any other window $w_l$ to be proportional to their overlap.
This makes the learning problem smoother and better behaved than when using a naive $0/1$ loss which equally penalizes all windows other than $l_I^*$.
Hence, we use the loss $\Delta(l,l') = 1 - \rho^\mathtt{overlap}(w_l,w_{l'})$ proposed by~\cite{blaschko08eccv}, which formalizes this intuition.
We can find $\bm{\alpha}$ by solving the following optimization problem 
\[
\min_{\bm{\alpha}, \xi} \frac{1}{2}||\bm{\alpha}||^2 + \gamma \sum_I(\xi_I)
\]
\begin{equation}
\textsc{s.t. }  \xi_I \geq 0,\forall I: \langle\bm{\alpha},\phi(X_I,l_I^*)\rangle - \langle \bm{\alpha},\phi(X_I,l) \rangle \geq \Delta(l,l_I^*) - \xi_I, \forall I, \forall l \in L_I \backslash l_I^*
\label{eq:str_svm}
\end{equation}
where $I$ indexes over all training images. This is a convex optimization problem, but it has hundred of thousands of constraints (i.e. about $1000$ candidate windows for each training image, times about $500$ training images per class in our experiments).
We solve it efficiently using quadratic programming with constraint generation~\cite{scholkopf2007predicting}. This involves finding the most violated constraint for a current $\bm{\alpha}$. We do this exactly as we can solve the inference problem (\ref{eq:the_loss}) and the loss $\Delta$ decomposes into a sum of terms which depend on a single window. 
Thanks to this, the constraint generation procedure will find the global optimum of (\ref{eq:str_svm})~\cite{scholkopf2007predicting}.

\seckiny
\section{Related work}
\label{sec:related_work}


The first work to try to annotate object locations in ImageNet~\cite{guillaumin12cvpr} addressed it as a window classification problem, where a classifier is trained on annotated images is then used to classify windows in images with no annotation. Later~\cite{vezhnevets14cvpr} proposed to regress the overlap of a window with the object using GP~\cite{RasmussenWilliams05book}, which allowed for self-assessment thanks to GP probabilistic output. We build on~\cite{vezhnevets14cvpr}, using Associative Embedding to learn the kernel between windows appearance features and GP to predict the spatial relations between a window and the object. Note how the model of~\cite{vezhnevets14cvpr} is equivalent to ours when using only the $\phi_S$ hyper-feature.

Importantly, both~\cite{guillaumin12cvpr,vezhnevets14cvpr}, as well as many other object localization techniques~\cite{wang13iccv, dalal05cvpr, girshick14cvpr, uijlings13ijcv, Vedaldi09}, score each window individually based only on its own appearance.
Our work goes beyond by evaluating windows based on richer cues measured outside the window. This is related to previous work on context~\cite{murphy03nips,heitz08eccv,choi10cvpr,desai:iccv09,rabinovich07iccv, harzallah09iccv,Torralba03} as well as to works that use structured output regression formulation~\cite{blaschko08eccv, Lampert08cvpr,vedaldi09nips,lehmann2011bmvc}. We review these areas below.

\parskiny \parskiny
\paragraph{Context.}

The seminal work of Torralba~\cite{Torralba03} has shown that global image descriptors such as GIST give a valuable cue about which classes might be present in an image (e.g. indoor scenes are likely to have TVs, but unlikely to have cars). 
Since then, many object detectors~\cite{murphy03nips,russell07nips,harzallah09iccv,uijlings13ijcv,ilsvrc13euuva,song11cvpr} have employed such global context to re-score their detections, thereby removing out-of-context false-positives. Some of these works also incorporate the region surrounding the object into the window classifier to leverage local context~\cite{dalal05cvpr,uijlings13ijcv,li11iccv,mottaghi14cvpr}.

Other works~\cite{rabinovich07iccv,heitz08eccv,choi10cvpr,desai:iccv09} model context as the interactions between multiple object classes in the same image.
Rabinovich et al.~\cite{rabinovich07iccv} use local detectors to first assign a class label to each segment in the image and then adjusts these labels by taking into account co-occurrence between classes.
Heitz and Koller~\cite{heitz08eccv} exploits context provided by "stuff" (background classes like road) to guide the localization of "things" (objects like cars).
Several works~\cite{choi10cvpr,desai:iccv09} model the co-occurrence and spatial relations between object classes in the training data and use them to post-process the output of individual object class detectors. 
An extensive empirical study of different context-based methods can found in~\cite{divvala09CVPR}.

The force driving those works is the semantic and spatial structure of scenes as arrangements of different object classes in particular positions.
Instead, our technique works on a different level, improving object localization for a {\em single class} by integrating cues from the appearance and spatial relations of all windows in an image. It can be seen as a new, complementary form of context.

\parskiny \parskiny
\paragraph{Localization with structured output regression} was first proposed by~\cite{blaschko08eccv}. They devised a training strategy that specifically optimizes localization accuracy, by taking into account the overlap of training image windows with the ground-truth. They try to learn a function which scores windows with high overlap with ground-truth higher than those with low overlap. The approach was extended by~\cite{vedaldi09nips} to include latent variables for handling multiple aspects of appearance and truncated object instances. At test time an efficient branch-and-bound algorithm is used to find the window with the maximum score. Branch-and-bound methods for localization where further explored in~\cite{Lampert08cvpr,lehmann2011bmvc}. 

Importantly, the scoring function in~\cite{blaschko08eccv,Lampert08cvpr,vedaldi09nips,lehmann2011bmvc} still scores each window in the test image independently. 
In our work instead we score each window in the context of all other windows in the image, taking into account their similarity in appearance space as well as their spatial relations in the image plane (sec.~\ref{sec:the_model}). We also use structured output regression~\cite{tsochantaridis:jmlr05} for learning the parameters of our scoring function (sec.~\ref{sec:str_svm}).
However, due to interaction between all windows in the test image, our maximization problem is more complex than in~\cite{blaschko08eccv,Lampert08cvpr}, making their branch-and-bound method inapplicable. Instead, we devise an early-rejection method that uses the particular structure of our scoring function to reduce the number of evaluations of its most expensive terms (sec.~\ref{sec:inference}).

%

\seckiny
\section{Experiments and conclusions}
\seckiny
\label{sec:experiments}
We perform experiments on the subset of ImageNet~\cite{deng09cvpr}  defined by~\cite{guillaumin12cvpr,vezhnevets14cvpr,guillaumin14ijcv}, which consists of 219 classes for a total of 92K images with ground-truth bounding-boxes. Following~\cite{vezhnevets14cvpr}, we split them in two disjoint subsets of 60K and 32K for training and testing respectively. The classes are very diverse and include animals as well as man-made objects (fig.~\ref{fig:Results}).
The task is to
localize the object of interest in images known to contain a given class~\cite{guillaumin12cvpr,vezhnevets14cvpr,guillaumin14ijcv}. 
We train a separate model for each class using the corresponding images from the training set.

\parskiny \parskiny
\paragraph{Features.}
\label{sec:implement_details}
For our method and all the baselines we use the same features as AE-GP+ method from~\cite{vezhnevets14cvpr}:
(i) three ultra-dense SIFT bag-of-words histograms on different color spaces~\cite{uijlings13ijcv} (each 36000 dimensions);
(ii) a SURF bag-of-word from~\cite{vezhnevets14cvpr} (17000 dimensions);
(iii) HOG~\cite{dalal05cvpr} (2048 dimensions).
%
%
We embed each feature type separately in a 10-dimensional AE~\cite{vezhnevets14cvpr} space. Next, we concatenate them together and add location and scale features as in~\cite{guillaumin12cvpr,vezhnevets14cvpr}.
In total, this leads to a 54-dimensional space on which the GP operates, i.e. only 54 parameters to learn for the GP kernel.

\seckiny
\subsection{Baselines and competitors} 

\paragraph{MKL-SVM.}
This represents a standard, classifier driven approach to object localization, similar to~\cite{Vedaldi09,uijlings13ijcv}.
On $90\%$ of the training set we train a separate SVM for each group of features described above.
%
We combine these classifiers by training a linear SVM over their outputs on the remaining $10\%$ of the data. We also include the location and scale features of a window in this second-level SVM.
This baseline uses exactly the same candidate windows~\cite{manen13iccv} and features as our method (sec.~\ref{sec:implement_details}).

\parskiny
\parskiny
\paragraph{UVA~\cite{uijlings13ijcv}.}
The popular method~\cite{uijlings13ijcv} can be seen as a smaller version of the MKL-SVM baseline we have just defined.
In order to make a more exact comparison to~\cite{uijlings13ijcv}, we remove the additional features and use only their three SIFT bag-of-words. Moreover, instead of training a second level SVM, we simply combine their outputs by averaging. This corresponds to~\cite{uijlings13ijcv}, but using the recent state-of-the-art object proposals~\cite{manen13iccv} instead of selective search proposals.
This method~\cite{uijlings13ijcv} is one of the best performing object detectors.
It has won the ILSVRC 2011~\cite{ilsvrc11} detection challenge and the PASCAL VOC 2012 detection challenge. 

\parskiny
\parskiny
\paragraph{AE-GP~\cite{vezhnevets14cvpr}.}
Finally, we compare to the AE-GP+ model of~\cite{vezhnevets14cvpr}. It corresponds to a degenerate version of our model which uses only $\phi_S$ to score each window in isolation by looking at its own features (\ref{eq:GP_term}). This uses the same candidate windows~\cite{manen13iccv} and features we use. 
This technique was shown to outperform earlier work on location annotation in ImageNet~\cite{guillaumin12cvpr}.

%

\begin{figure*}
\begin{center}
\includegraphics[scale=0.4]{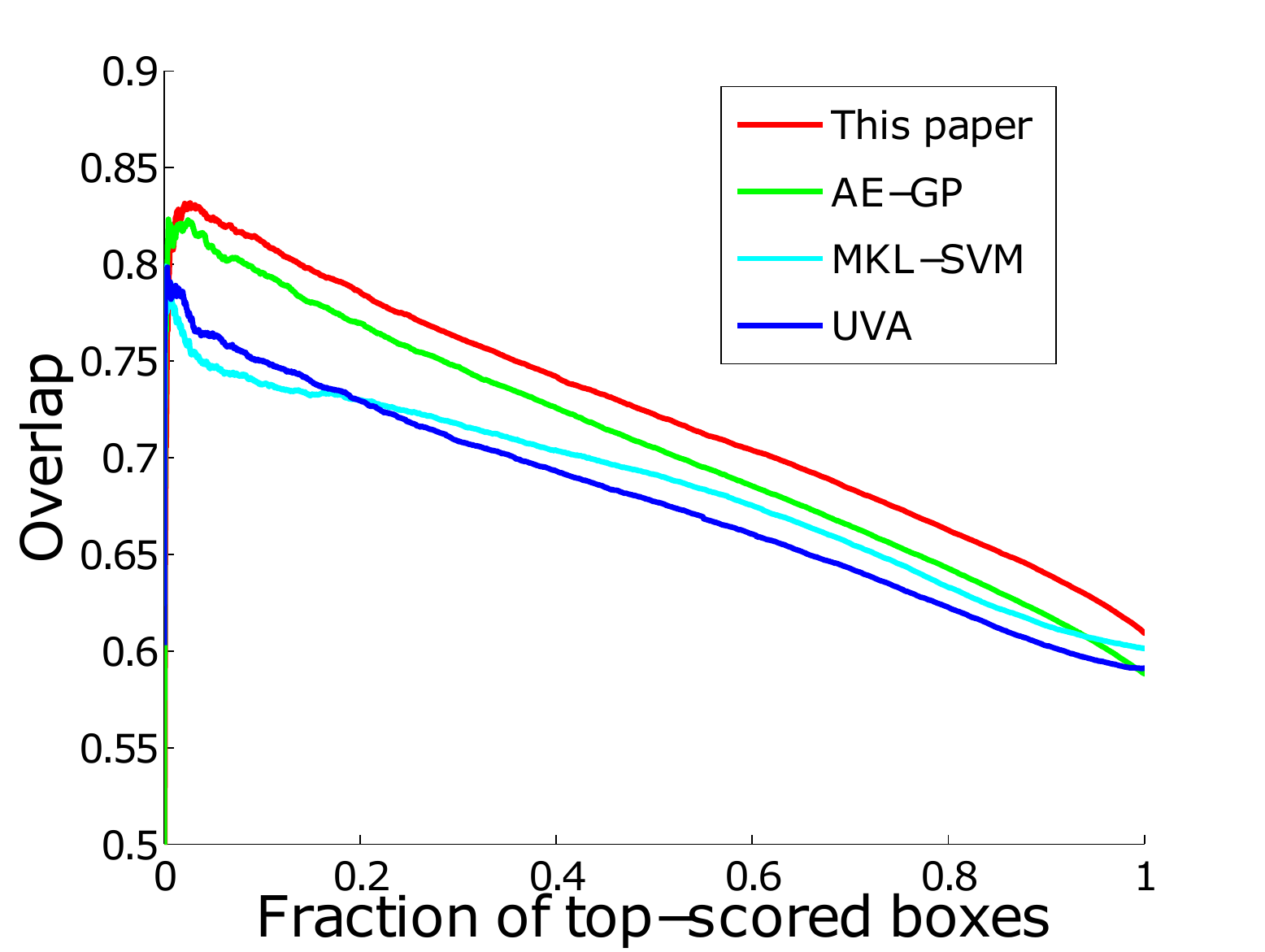}           
\end{center}
\figskiny 
\caption{\small{\it   
{\bf Mean overlap curves.} \label{subfig:self_asses} We retain the single top scoring candidate window in a test image and measure the mean overlap of the output windows with the ground-truth. We vary a threshold on the score of the output windows to generate performance curves. The higher the curve, the better.
}}
\figskiny   
\end{figure*}

\seckiny
\subsection{Results}
\seckiny
For each method we retain the single top scoring candidate window in a test image.
We measure localization accuracy as the mean overlap of the output windows with the ground-truth~\cite{vezhnevets14cvpr} (fig.~\ref{subfig:self_asses}).
We vary a threshold on the score of the output windows to generate performance curves. The higher the curve, the better.

\begin{figure*}
\begin{center}
    \includegraphics[scale=0.195] {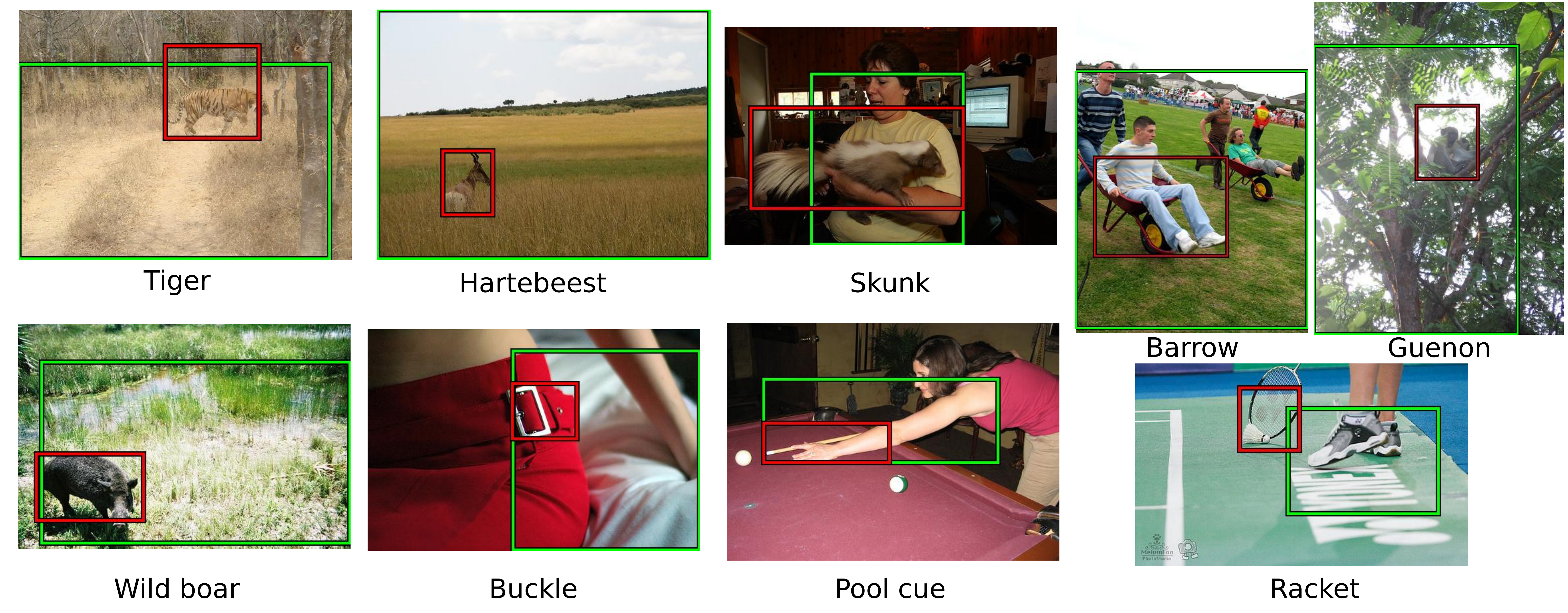}  
\end{center}
\figskiny 
\caption{\small{\it   
{\bf Qualitative results.} Results of our method (red) vs AE-GP~\cite{vezhnevets14cvpr} (green).} Notice, how our method is able to detect small, off-center objects despite occlusion (pool cue) or the object blending with its surroundings (tiger). \label{fig:Results} }
\figskiny  \figskiny  
\end{figure*}

As fig.~\ref{subfig:self_asses} shows, the proposed method consistently outperforms the competitors and the baseline over the whole range of the curve.
Our method achieves 0.75 mean overlap when returning annotations for 35\% of all images. At the same accuracy level, AE-GP, MKL-SVM and UVA return $28\%$, $9\%$ and $4\%$ images respectively, i.e. we can return $7\%$ more annotation at this high level of overlap than the closest competitor.
Producing very accurate bounding-box annotations is important for their intended use as training data for various models and tasks.
%
Improving over AE-GP validates the proposed idea of scoring candidate windows by taking into account spatial relations to all other windows and their appearance similarity.
%
%
The favourable comparison to the excellent method UVA~\cite{uijlings13ijcv} and to its extended MKL-SVM version demonstrates that our system offers competitive performance.
Example objects localized by our method and by AE-GP are shown in fig.~\ref{fig:Results}.
Our method successfully operates in cluttered images (guenon, barrow, skunk). It can find camouflaged animals (tiger), small objects (buckle, racket), and deal with diverse classes and high intra-class variation (pool cue, buckle, racket).

To evaluate the impact of our fast inference algorithm (sec.~\ref{sec:inference}) we compared it to brute force (i.e. evaluating the energy for all possible configurations) on the baseball class. On average brute force takes 17.6s per image, whereas our fast inference takes 0.14s. Since our inference method is exact, it produces the same solution as brute force, but 124$\times$ faster.


%

\seckiny
\subsection{Conclusion}
\seckiny
We have presented a new method for annotating the location of objects in ImageNet, which goes beyond considering one candidate window at a time. Instead, it scores each window in the context of all other windows in the image, taking into account their similarity in appearance space as well as their spatial relations in the image plane.
As we have demonstrated on 92K images from ImageNet, our method improves over some of the best performing object localization techniques~\cite{uijlings13ijcv,vezhnevets14cvpr}, including the one we build on~\cite{vezhnevets14cvpr}.
\seckiny
\paragraph{Acknowledgment} This work was supported by the ERC Starting Grant VisCul. A. Vezhnevets was also supported by SNSF fellowship PBEZP-2142889.

\seckiny
\bibliography{shortstrings,calvin,vggroup}
\end{document}